# Analyzing Consumer Reviews for Understanding Drivers of Hotels' Ratings: An Indian Perspective

Subhasis Dasgupta
Department of Data Science
Praxis Business School
Kolkata, INDIA
email: subhasisdasgupta1@acm.org

Soumya Roy
Department of Data Science
Praxis Business School
Kolkata, INDIA
email: soumya.roy@praxis.ac.in

Jaydip Sen
Department of Data Science
Praxis Business School
Kolkata, INDIA
email: jaydip.sen@acm.org

*Abstract*— In the internet era, almost every business entity is trying to have its digital footprint in digital media and other social media platforms. For these entities, word of mouse is also very important. Particularly, this is quite crucial for the hospitality sector dealing with hotels, restaurants etc. Consumers do read other consumers' reviews before making final decisions. This is where it becomes very important to understand which aspects are affecting most in the minds of the consumers while giving their ratings. The current study focuses on the consumer reviews of Indian hotels to extract aspects important for final ratings. The study involves gathering data using web scraping methods, analyzing the texts using Latent Dirichlet Allocation for topic extraction and sentiment analysis for aspect-specific sentiment mapping. Finally, it incorporates Random Forest to understand the importance of the aspects in predicting the final rating of a user.

## I. INTRODUCTION

The dynamics of businesses are changing and only those organizations will flourish which can adapt the change and can move along. This is particularly important in service sector because the element of trust play an important role in final buying intention [1]. It is now well known that consumers do read reviews of the product before making the final decision to either purchase or not to purchase. Particularly, when the associated price is high, then these reviews can have significant impact on the purchase decision. In hospitality sector, the hygiene factor may be one such factor where the consumers would not like to compromise. And, hotel businesses are very much aware of this factor. However, apart from this, service sectors also have three extra Ps such as People, Process and Physical Evidence. Any lapses in these areas are also going to have negative impact on the business. This is the reason; hotel industry pays special focus on maintaining the service levels in these areas. In fact, hotels which are maintaining high standards in these areas can charge premiums for their services and satisfied consumers do pay the premiums for the extra experiences. Since hoteling business contribute significantly in the GDP of India [2], it is also important to understand which factors play important roles in deriving satisfactions in the minds of the consumers.

When it comes to consumer reviews, the sheer volume sometimes acts as a deterrent for both the consumers and also to the management of the hotels. Going through the reviews to understand the pain points or the point of appreciations is a laborious task for the management. Since the study is closely connected to the management of the hotels, the focus is kept to help the hotel managements to understand the reviews and evaluate them objectively using the existing tools in dealing with texts. There are several tools available to deal with texts and in this study, the most commonly used tools were used to focus more on the business problem rather than the technicalities of the tools. The study does not involve usage of any sophisticated AI tools and yet it could extract meaningful outcomes from business point of view.

The paper is structured as follows: Section II provides a brief overview of significant recent studies on review analysis. Section III discusses the data extraction process and outlines the step-by-step research methodology. Section IV details the results of the structured analysis. Finally, Section V offers the paper's conclusions..

## II. RELATED WORK

Analyzing consumer reviews is not new and there are several prior works available in this research domain. However, the important part is that the researchers used various tools to extract different types of information. Particularly, in the domain of hotel industry there are quite a few works such as [3][4][5]. Not only hotel reviews, there are also works available on mobile phone reviews [6] which was used to detect market structure of mobile phone brands in India. In fact, the researcher could use topic modeling [7] to analyze the reviews on phones in a structured manner [8]. This is where the idea of extending the work for hotel industry was derived. Using text mining and social media analysis in hotel reviews is not new, however. Wang et al, investigated the emotional characteristics and thematic content of star-rated hotels in Nanchang, China, through the lens of social sensing [9]. The authors use text mining and sentiment analysis on online reviews from social media platforms to gather data on customer emotions and experiences. The study identifies key themes such as service quality, cleanliness, and amenities that significantly influence hotel ratings and customer satisfaction. The analysis reveals that positive reviews often highlight aspects like excellent service and comfortable accommodations, while negative reviews frequently mention issues like poor hygiene and adequate facilities. The insights gained from this work provide actionable recommendations for hotel managers to improve service offerings and enhance overall guest experiences based on real-time feedback from social media.

Zvarevashe & Olugbara propose a framework that uses natural language processing techniques for processing hotel reviews, including tokenization, stop-word removal, and stemming/lemmatization [10]. Feature extraction methods like TF-IDF and word embeddings such as Word2Vec and GloVe are used to represent text data. Sentiment classification is performed using machine learning algorithms such as Naïve Bayes, support vector machine (SVM), and random forests, and deep learning models, such as recurrent neural networks (RNN) and convolutional neural networks (CNN). The results demonstrate that deep learning models with word embeddings excel in sentiment classification, providing detailed insights into customer feedback.

Wen et al. evaluate the performance of BERT and ERNIW models in sentiment analysis of Chinese hotel reviews. Preprocessing steps include tokenization, special character removal, and text normalization [11]. Both models are fine-tuned on the hotel review dataset, leveraging their transformer architectures to classify reviews into positive, negative, and neutral categories. Evaluation metrics such as accuracy, precision, recall, and F1-score show that both models perform well, with ERNIE slightly outperforming BERT due to its integration of external knowledge.

Topic modeling does a good job when the volume of data is more and the texts are not having too many rare words such as mixture of English and a local language dialect. Topic modeling is an unsupervised learning methodology making it quite applicable for analyzing reviews written in English. Another important aspect of topic modeling is that it does automatic document clustering based on the topics. Since the number of reviews available is quite large, application of topic modeling was found appropriate in the current study to extract topics in an unsupervised manner.

Extraction of topics is one important aspect but it was also necessary to find the sentiment scores associated with the documents under consideration. Hence, established tools were required to perform sentiment analysis. There are quite a few tools available for sentiment analysis such as lexicon based sentiment analysis used in VADER [12] and other deep learning based models such as BERT [13]. Since BERT model takes into account of contexts of the texts, it often performs better way than pure lexicon based models. Hence, in the current study, BERT based model was considered for extracting sentiment scores.

Finally, topic specific sentiment scores were needed for building classification model to understand the importance of topics in users' ratings. For this purpose, different classification models were decided to be used within the study and choose the model with best performance. A few models under consideration were RandomForest, Logistic Regression, XgBoost and LightGBM.

III. DATA AND METHODOLOGY

The methodology consists of a few steps which are explained below with respective sub subsections.

*(1) Data extraction:* The data were extracted using web scraping process using the python package Beautiful Soup. For this purpose, Trip Advisor website was chosen and publicly available reviews were scraped. While selecting the hotels, it was kept in mind that those hotels are chosen which are distributed in different regions of India. Intentionally, a few hotels were included which are less popular or less visited so that sufficient number of reviews can be got with lower user ratings. In total, 186 different hotels were identified from 20 different states of India to get a fair coverage of the market. The total number of reviews collected was 44363 with user ratings. While extracting the data, care was taken so that the website never got overloaded due to frequent posting of quick queries.

*(2) Data Preprocessing:* While dealing with textual data, an important step is preprocessing of the dataset. Data preprocessing involved removal of unnecessary punctuations, removal of numbers, removal of rare words, lemmatization of the words for extracting root word and finally creating the final Bag of Words collection from the textual data. For creating the bag of words, python package spacy was used which produced 273999 token from the 44363 reviews. The texts were broken down to their respective sentences. The sentences had several stopwords which had very low classification powers (e.g., "a", "the", "about" etc.). Those stopwords were also removed prior to using the tokens for the next task, that is, topic modeling. Finally, texts were stored as list of list of tokenized words after preprocessing.

IV. ANALYSIS AND RESULTS

Once data preprocessing was done, the next task was extracting topics from the texts using topic modeling. Extracting topics in an unsupervised manner is the objective of topic modeling. However, this extraction needs to be done carefully. For topic modeling, genism package was used. Topic modeling involves two Dirchlet parameters α and η. Apart from these, the number of topics to be extracted is also needed to be fixed before training the model. These hyperparameters have a prominent impact on the final performance of the model. Hence, they were required to be fixed after optimization. The metric used in this process was coherence score. There is another metric available for this purpose called perplexity. However, compared to perplexity, coherence score was found to be more meaningful in interpreting the topics [14][15]. Hence, coherence score was used to optimize the hyperparameters in topic modeling. Since the Dirichlet parameters can take any floating point value between [0, ∞), instead of Grid Search or Random Search, Bayesian Search was used. Unlike Grid or Random search, Bayesian search retains the outcomes of previous searches to progressively produce better set of hyperparameters for the task in hand. More importantly, in Bayesian Search, the sample space is infinite in theory which is not the case in Grid or Random Search. Thus, a Bayesian Search was considered most appropriate in the given context. Particularly in Bayesian Search, Tree structured Parzen Estimator (TPE) method was used for its simplicity and versatility. Total 20 iterations were run and the best solution provided by the coherence score was 0.65 with number of topics as 7 and α and η values being 0.149 and 0.501 respectively. A coherence score above 0.55 is usually considered good and the best result found was having a score 0.65. Hence, the topic model was rerun with these optimized hyperparameter values. The topics along with the most important words in each topic is shown in Table 1.

**Table 1.** Top 10 most important words in each topic

| Topic 1 | Topic 2 | Topic 3 | Topic 4 | Topic 5 | Topic 6 | Topic 7 |
|---|---|---|---|---|---|---|
| pool | good | staff | room | mr | hotel | food |
| resort | room | service | hotel | team | stay | restaurant |
| area | service | helpful | one | u | visit | breakfast |
| view | food | friendly | u | thanks | stayed | good |
| beach | nice | always | time | special | experience | buffet |
| beautiful | hotel | care | get | chef | place | dinner |
| activity | excellent | courteous | even | thank | best | delicious |
| spa | well | really | day | front | time | spread |
| place | great | u | could | mention | taj | option |
| swimming | clean | make | star | made | would | served |

**Table 2**: Topic coherence scores

| Topics | Topic 1 | Topic 2 | Topic 3 | Topic 4 | Topic 5 | Topic 6 | Topic 7 |
|---|---|---|---|---|---|---|---|
| Scores | 0.6739 | 0.5236 | 0.6043 | 0.7048 | 0.7517 | 0.5726 | 0.7416 |

The topic wise coherence scores are given in Table 2. The coherence scores of topic 2 and topic 6 are slightly lesser than 0.55. The exact identification of topics is sometimes difficult from the top n terms. Hence, each sentence was passed through the topic model to understand the topic associated with that sentence. Afterward, for a particular topic, the sentences were manually analyzed. From the analysis, it was understood that 'Topic 1' was dealing with mostly the activities the customers did while staying in a hotel. In some cases, some customers were seeing complaining about lack of provision for any activity. Hence, topic 1 was identified as 'scope of activity'. Topic 2 was dealing with the location of the hotel as the sentences associated with this topic mentioned clearly about the advantages or disadvantages of the location of the hotel. Hence, topic 2 was identified as 'location'. In topic 3, the customers were found to discuss about the behavior of the hotel staffs. Hence, this topic was identified as 'hotel staff'. In topic 4, evaluation of hotel rooms got the highest importance. The sentences associated with topic 4 where mentioning about the experience with hotel rooms. Hence, this topic was identified as 'quality of rooms'. The 5[th] topic is a little bit interesting one as in this topic, the sentences were found to mention about the role of the hotel managers and, sometimes, some other person while dealing with the issue or event. Hence, this topic was identified as 'special personnel'. Topic 6 was found to contain a mixture of different aspects such as 'family visit', 'modern décor', 'nights of stay' etc. Hence, this topic was named as 'miscellaneous' since it was not mentioning any specific aspect in general. The last topic, that is, topic 7 was found to deal with food quality. The sentences associated with topic 7 was mentioning various food items, their quality and tastes. Hence, this topic was identified as 'food quality'. The final mapping is also shown in Table 3. While doing topic modeling, each review was broken down to the respective sentences and only after that topic modeling was done after text preprocessing. Hence, the number of sentences reached 2.4 lacs. There were some sentences which were carrying emoji only and they could be put inside any topic. For those sentences, the topic allocation was NaN, that is, missing topics.

**Table 3**: Identified topics in topic modeling

| Topics  | Identified aspect  |
|---------|--------------------|
| Topic 1 | Scope of Activity  |
| Topic 2 | Location           |
| Topic 3 | Hotel staffs       |
| Topic 4 | Quality of rooms   |
| Topic 5 | Special personnel  |
| Topic 6 | Miscellaneous      |
| Topic 7 | Food quality       |

The next task involved was extracting sentence sentiment analysis. For this purpose, deep learning based model was used for more accurate sentiment mapping. For this purpose, pre-trained DistilBERT [16] model was chosen as this is a lighter version of the BERT model with comparable performance. The pre-trained models are hosted in huggingface.com [1] and there are several text classifiers available for multiple classes. But, for this study, that model was chosen which was finetuned with SST-2-English corpus as that model was used more frequently for sentiment classification by the research community. As on date, there were more than 6 million downloads of the model by various researchers and practitioners. However, DistilBERT does text classification with only two classes, that is, POSITIVE and NEGATIVE and it also provides the probability of the classes. The objective of the study was to find out the influence of the aspects on user ratings in an objective manner. Hence, it was decided that instead of the binary values, continuous values of sentiment scores would be used. For this purpose, probability of POSITIVE class was considered as the probability of success (p) and probability of NEGATIVE class was considered as the probability of failure (1-p). In binary classification, the probability of success is defined as:

$$p = \frac{\exp[f(x)]}{1 + \exp[f(x)]} \quad (1)$$

where, f(x) is the required regression score which can be easily considered as the sentiment score for the text. After rearrangements, the regression score can be calculated as:

$$f(x) = \ln\left(\frac{p}{1-p}\right) \quad (2)$$

If the sentiment label was POSITIVE, the value of f(x) remains positive and if the sentiment label was predicted as NEGATIVE, the regression score was multiplied with -1 to get a negative value. This way the sentiment scores were generated for each sentence within the dataset. Since, for each review, there were multiple sentences and each sentence was mapped with topic as well as the sentiment scores, the next step involved defining each review as a collection of topics with sentiment scores. A sample output is given in Table 4.

**Table 4**: Sample data showing conversion of review texts as collection of topics with sentiment score

| Review ID | Topic 1 | Topic 2 | Topic 3 | Topic 4 | Topic 5 | Topic 6 | Topic 7 | Ratings |
|-----------|---------|---------|---------|---------|---------|---------|---------|---------|
| 1 | 7.452 | 0.000 | 8.938 | 2.891 | 0.000 | 6.570 | 0.000 | 5 |
| 2 | 8.872 | 8.971 | 7.955 | 0.000 | 0.000 | 7.577 | -4.154 | 5 |
| 3 | 0.000 | 9.042 | 3.378 | -5.741 | 2.955 | 2.530 | 8.280 | 5 |
| 4 | 0.000 | 0.000 | 8.919 | 0.826 | 0.000 | 8.999 | 0.000 | 5 |
| 5 | 8.019 | 8.916 | 8.759 | 0.000 | 8.259 | 5.327 | 0.000 | 5 |

As per Table 4, the first review contains 4 topics and the rating provided by the reviewer is 5 out of 5. Topics with 0 sentiment score means that the review does not contain the specific topic and hence the reviewer is neutral to that topic. The final dataset created in this way contained 44359 rows and 9 columns.

To understand the influence of the topic sentiments on the user ratings, classification modeling was considered since the ratings were discrete with fixed values, i.e., {1,2,3,4,5}. To start with multiclass classification with 5 classes were considered. Regularized logistic regression with deviance loss was considered as the first model since logistic regression falls under linear model and linear models are very good for interpretation of the results. The distributions of the classes were quite imbalanced in nature with almost 64.47% of the ratings were 5 and just 3.95% ratings were 2. Around 10% of the reviewers gave their ratings as 3. With such imbalance, accuracy was not a good metric to evaluate the performance of the model. Hence, Cohen's Kappa score was considered as the preferred matrix because kappa score takes into account the chance factor which makes the metric better than the accuracy metric for imbalanced dataset. The respective kappa scores for 4 different popular algorithms are shown in Table 5.

---

[1] https://huggingface.co/docs/transformers/model_doc/distilbert

**Table 5:** Cross validation scores of modeling algorithms for 5 class classification

| Model | Cohen's Kappa Score (3 fold cross validation) |
|---|---|
| Logistic Regression | 0.3420 |
| Random Forest | 0.4735 |
| **Xgboost** | **0.4761** |
| LightGBM | 0.4727 |

The kappa score of Xgboost was found to be the best and the same for Logistic Regression was the worst. This clearly suggested that the classes were not linearly separable and Xgboost found to perform better than Random Forest marginally. A kappa score of 0.47 suggest fair level of agreement between the actual classes and the predicted classes. Hence, a different strategy was adopted. From the 5 classes, 3 classes were created, positive, neutral and negative. The rating 3 was considered as neutral, ratings 4 and 5 were considered positive and ratings 1 and 2 were considered negative. With this new classes, the same models were retrained. With 3 classes, instead of 5 classes, the kappa score improved and the new kappa scores are shown in Table 6. A kappa score reaching close to 0.6 can be considered as moderate level of agreement and the best kappa score was again obtained by Xgboost.

**Table 6:** Cross validation scores of modeling algorithms for 3 class classification

| Model | Cohen's Kappa Score (3 fold cross validation) |
|---|---|
| Logistic Regression | 0.5003 |
| Random Forest | 0.5869 |
| **Xgboost** | **0.5882** |
| LightGBM | 0.5872 |

Even though Xgboost is a highly non-linear modeling technique, it can provide its own feature importance based on the 'gain' measure. The feature which could provide the maximum reduction of impurity attains the maximum gain score. Figure 1 shows the feature importance based on this gain score.

According to Figure 1, for overall performance in classification task, the most important feature is "Quality of rooms". However, this is a multiclass classification problem and hence, the overall picture is not very informative to understand which features are important for a particular class prediction. Because of this, shapley values were considered to understand the importance of features for a particular class.

Since there were three classes, three such importance plots were generated as shown in Figure 2, Figure 3 and Figure 4.

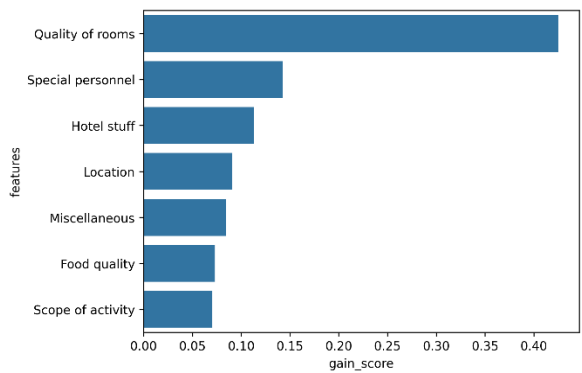

**Figure 1**: Overall feature importance as per Xgboost model (using gain score)

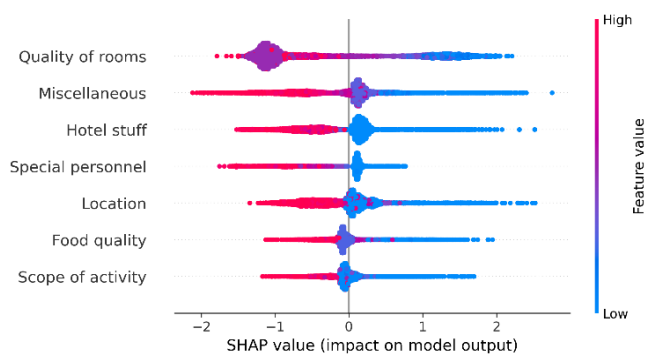

**Figure 2**: Feature importance plot for class 'Negative'

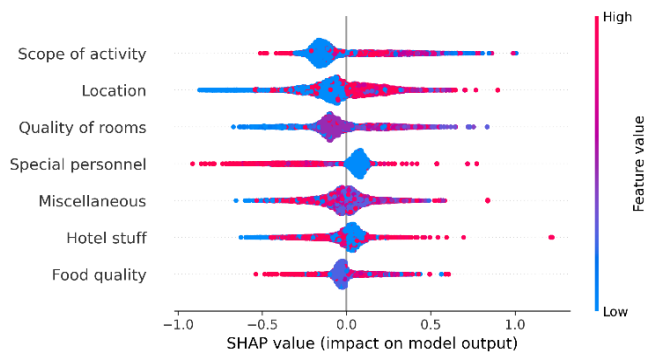

**Figure 3**: Feature importance plot for class 'Neutral'

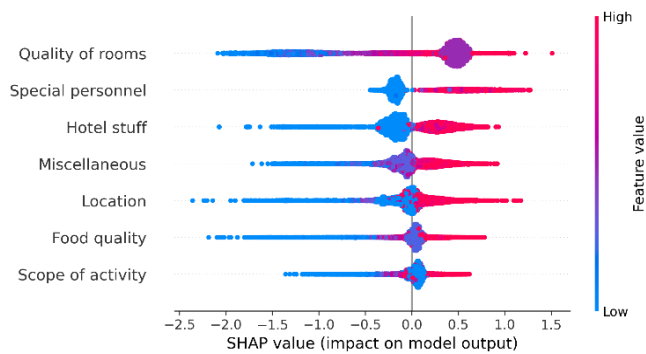

**Figure 4**: Feature importance plot for class 'Positive'

As the SHAP value increases, the probability towards the targeted class also increases. Figure 2 shows that as the sentiment score increases (accordingly the SHAP value will also increase), the probability towards the 'Negative' class will decrease. In fact, this is quite evident because all the features were handling the sentiment scores and if the sentiment score increases, the rating should also increase. Also, if both Figure 2 and Figure 4 are seen, the orders of importance of features are very similar for the two classes except for two topics, that is, 'Special personnel' and 'Miscellaneous'. The importance of these features got swapped in these two figures. But, if Figure 3 is looked at, it can be seen that the order of feature importance got changed. Interestingly, 'Scope of activity' got highest preference for this class under consideration, i.e., Neutral. This class is rather difficult to predict compared to the other two classes. This is because as the sentiment scores start going higher, the class will shift to 'Positive' class and if the sentiment score starts going down, it would shift to the 'Negative' class. That is why, the plot contains mixtures of high and low values in the dot plot unlike the cases in Figure 2 and Figure 4 where the red dots and blue dots are quite separated. This also gives an idea of the expectations of the customers who gave 3 ratings to the hotel they stayed in. It not conclusive but it seems that these customers have some latent need of recreations while staying in a hotel. To test this hypothesis, further deep down analysis would be required with statistical tests which is beyond the scope of the current study.

## V. CONCLUSION

In this paper, more than 44 thousand consumer reviews were analyzed. The reviews were all pertaining to Indian hotels and the total number of hotels was also 186. The researchers in this study have demonstrated that by combining machine learning model such as Xgboost with models associated with textual data analysis such as topic modeling, interesting insights can be extracted from the textual data without thorough manual reading and comprehension. In this work, quite a few concepts were used such as latent dirichlet allocation method for topic extraction, analyzing coherence score for meaningful topic evaluation, DistilBERT based sentiment analysis and finally Xgboost for building classification model to predict user rating. For visualizing the topic importance, SHAP values were considered for model interpretation. This way, a comprehensive analysis was done.

## VI. REFERENCES

[1] C. X. Ou and C. L. Sia, "To trust or to distrust, that is the question: Investigating the trust-distrust paradox," *Commun. ACM*, vol. 52, no. 5, pp. 135–139, 2009.

[2] E. Times, "Hotel industry's contribution to India's GDP to hit $1 trillion by 2047: HAI," *The Economic Times*, Aug. 20, 2023. [Online]. Available: https://economictimes.indiatimes.com/industry/services/hotels-/-restaurants/hotel-industrys-contribution-to-indias-gdp-to-hit-1-trillion-by-2047-hai/articleshow/102870459.cms?from=mdr

[3] A. García-Pablos, M. Cuadros, and M. T. Linaza, "Automatic analysis of textual hotel reviews," *Inf. Technol. Tour.*, vol. 16, no. 1, pp. 45–69, 2016.

[4] K. Zvarevashe and O. O. Olugbara, "A framework for sentiment analysis with opinion mining of hotel reviews," in *2018 Conference on information communications technology and society (ICTAS)*, IEEE, 2018, pp. 1–4.

[5] V. Chang, L. Liu, Q. Xu, T. Li, and C. Hsu, "An improved model for sentiment analysis on luxury hotel review," *Expert Syst.*, vol. 40, no. 2, p. e12580, 2023.

[6] S. Dasgupta and K. Sengupta, "Consumer Reviews for Market Structure Analysis: A Text Mining Approach," *LBS J. Manag. Res.*, vol. 13, no. 1, pp. 44–53, 2015.

[7] D. M. Blei, A. Y. Ng, and M. I. Jordan, "Latent dirichlet allocation," *J. Mach. Learn. Res.*, vol. 3, no. 3, pp. 993–1022, 2003.

[8] S. Dasgupta and K. Sengupta, "Consumer Review Analysis Using Topic Modelling," *LBS J. Manag. Res.*, vol. 15, no. 1, pp. 48–57, 2017.

[9] J. Wang, Y. Xia, and Y. Wu, "Emotional characteristics and theme mining of star-rated hotels from the perspective of social sensing: A case study of Nanchng City, China, Computational Uran Science, Vol 2, No 10, 2022.

[10] K. Zvarevashe and O. O. Olugbara, "A framework for sentiment analysis with opinion mining of totel reviews", in Proceedings of the 2018 Conference on Information Communications Technology and Society (ICTAS), Durban, South Africa, 2018, pp 1-4.

[11] Y. Wen, Y. Liang, and X. Zhu, "Sentiment analysis of hotel online reviews using BERT model and ERNIE model - Data from China", PLoS ONE, Vol 8, No 3, 2023.

[12] C. Hutto and E. Gilbert, "Vader: A parsimonious rule-based model for sentiment analysis of social media text," in *Proceedings of the international AAAI conference on web and social media*, 2014, pp. 216–225.

[13] J. Devlin, M.-W. Chang, K. Lee, and K. Toutanova, "Bert: Pre-training of deep bidirectional transformers for language understanding," *arXiv Prepr. arXiv1810.04805*, 2018.

[14] D. Newman, J. H. Lau, K. Grieser, and T. Baldwin, "Automatic evaluation of topic coherence," in *Human language technologies: The 2010 annual conference of the North American chapter of the association for computational linguistics*, 2010, pp. 100–108.

[15] D. Mimno, H. Wallach, E. Talley, M. Leenders, and A. McCallum, "Optimizing semantic coherence in topic models," in *Proceedings of the 2011 conference on empirical methods in natural language processing*, 2011, pp. 262–272.

[16] V. Sanh, L. Debut, J. Chaumond, and T. Wolf, "DistilBERT, a distilled version of BERT: smaller, faster, cheaper and lighter," *arXiv Prepr. arXiv1910.01108*, 2019.